\documentclass[conference]{IEEEtran}
\usepackage{cite}
\usepackage{amsmath,amssymb,amsfonts}
\usepackage{algorithmic}
\usepackage{graphicx}
\usepackage{textcomp}
\usepackage{xcolor}
\def\BibTeX{{\rm B\kern-.05em{\sc i\kern-.025em b}\kern-.08em
    T\kern-.1667em\lower.7ex\hbox{E}\kern-.125emX}}
\begin{document}

\title{Few-shot learning approaches for classifying low resource domain specific software requirements\\
}

\author{\IEEEauthorblockN{Anmol Nayak}
\IEEEauthorblockA{\textit{ARiSE Labs at Bosch}\\
Bengaluru, India \\
Anmol.Nayak@in.bosch.com}
\and
\IEEEauthorblockN{Hari Prasad Timmapathini}
\IEEEauthorblockA{\textit{ARiSE Labs at Bosch}\\
Bengaluru, India \\
HariPrasad.Timmapathini@in.bosch.com}
\and
\IEEEauthorblockN{Vidhya Murali}
\IEEEauthorblockA{\textit{ARiSE Labs at Bosch}\\
Bengaluru, India \\
Vidhya.Murali@in.bosch.com}
\and
\IEEEauthorblockN{Atul Anil Gohad}
\IEEEauthorblockA{\textit{ARiSE Labs at Bosch}\\
Bengaluru, India \\
AtulAnil.Gohad@in.bosch.com}
}


\maketitle

\begin{abstract}
With the advent of strong pre-trained natural language processing models like BERT, DeBERTa, MiniLM, T5, the data requirement for industries to fine-tune these models to their niche use cases has drastically reduced (typically to a few hundred annotated samples for achieving a reasonable performance). However, the availability of even a few hundred annotated samples may not always be guaranteed in low resource domains like automotive, which often limits the usage of such deep learning models in an industrial setting. In this paper we aim to address the challenge of fine-tuning such pre-trained models with only a few annotated samples, also known as Few-shot learning. Our experiments focus on evaluating the performance of a diverse set of algorithms and methodologies to achieve the task of classifying BOSCH automotive domain textual software requirements into 3 categories, while utilizing only 15 annotated samples per category for fine-tuning. We find that while SciBERT and DeBERTa based models tend to be the most accurate at 15 training samples, their performance improvement scales minimally as the number of annotated samples is increased to 50 in comparison to Siamese and T5 based models.
\end{abstract}


\section{Introduction}
Few-shot learning (FSL) is a crucial area of research in machine learning that focuses on developing algorithms that can learn to recognize patterns with very limited training data. The aim of FSL is to equip machines with the ability to generalize from limited experiences, similar to the way human beings can recognize new patterns with only a few examples. The challenge of FSL lies in the fact that deep neural networks, which are commonly used in computer vision and natural language processing (NLP), require large amounts of training data to perform well. This has motivated the development of innovative approaches that can effectively learn from few examples and generalize to new, unseen data. The exact number of examples considered "few" can vary depending on the task and the complexity of the model. In general, FSL in NLP is often considered to involve learning from less than a hundred examples per class.

There have been many research papers in recent years that apply FSL in the field of NLP. These papers aim to address the challenge of learning from limited data in NLP tasks such as text classification, named entity recognition, and machine translation. One common approach in these papers is to use meta-learning, where the model is trained on a variety of tasks with limited data and then fine-tuned on a specific task with few examples. Another approach is to use transfer learning, where pre-trained language models are fine-tuned on a few examples of the target task. A comprehensive survey of these papers showed the effectiveness of FSL in NLP and provides insights into the design of models and algorithms that can perform well with limited data \cite{b1}.

\begin{figure*}[htbp]
\centerline{\includegraphics[scale=0.58]{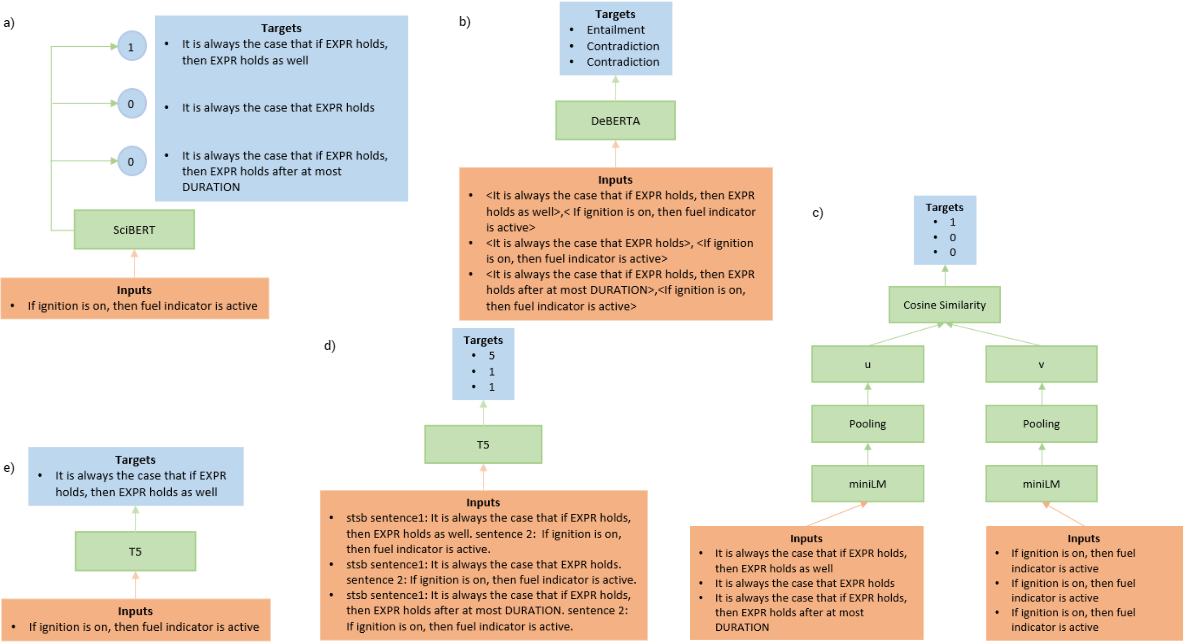}}
\caption{Architecture overview of the proposed approaches: a) SciBERT encoder + Linear Classification head b) DeBERTa encoder + Natural Language Inference head c) Siamese network with MiniLM encoder + Semantic Textual Similarity head d) T5 encoder + Semantic Textual Similarity target e) T5 encoder + Classification Generative Text target.}
\label{fig}
\end{figure*}

Further, there have been several research papers that have explored the use of data augmentation techniques for FSL in NLP. Data augmentation involves generating synthetic data from existing data, which can be used to increase the size of the training set and improve model performance. A comprehensive survey of these techniques showed that this can be achieved with rule based techniques like replacing named entities with synonyms or similar expressions, model based techniques like back translation or generative approaches \cite{b2}.

Recent advancements in pre-trained transformer models such as BERT \cite{b3}, DeBERTa \cite{b4}, RoBERTa \cite{b5}, MiniLM \cite{b6}, T5 \cite{b7} have lead to state-of-the art performance on several NLP tasks of the General Language Understanding Evaluation (GLUE) benchmark \cite{b8}. This has enabled the use of such models for various downstream tasks with minimal domain specific fine-tuning. However, domain specific annotated data is often very limited and difficult to generate in low resource domains like automotive as it requires a domain experts understanding to perform annotation. While fine-tuning tends to give good performance when the number of annotated training samples are in hundreds or more, the performance of such pre-trained models falls drastically when the number of annotated samples are very few due to the following reasons:

\begin{itemize}
    \item Difficulty in learning the semantics of a niche domain as they are significantly different from the pre-training data that was used to train these models.
    \item Complexity and Uniqueness of the domain specific tasks in comparison to the pre-training tasks of these models.
    
\end{itemize}

NLP for Requirements Engineering is a growing field with many recent works focusing on requirements classification by fine-tuning pre-trained models \cite{b9,b10,b11}. Further, there have been a few works that focus on building NLP approaches for domain specific language use cases \cite{b12,b13,b14,b15,b16,b17}. Although still a nascent field, there have been a few works in using FSL for requirements classification \cite{b18,b19}. While these prior works provide great insights into using pre-trained models for a requirements engineering context, our work builds upon them in the following aspects:

\begin{itemize}
    \item Larger test set ($\sim$3000 BOSCH requirements) for validating the efficacy of different approaches with a greater confidence.
    \item Reformulating the classification task as 5 different NLP tasks using recent state-of-the-art architectures.
    \item Strictly constraining the training set to a few samples.
    
\end{itemize}

\begin{table*}[htbp]
\caption{Hyper parameter settings for the different algorithms}
\begin{center}
\begin{tabular}{|c|c|c|c|c|c|}
\hline
\textbf{Algorithm}&\multicolumn{5}{|c|}{\textbf{Hyper parameter}} \\
\cline{2-6} 
\textbf{} & \textbf{\textit{Version}}& \textbf{\textit{Epochs}}& \textbf{\textit{Optimizer}}& \textbf{\textit{Learning Rate}} & \textbf{\textit{Learning Rate Warm-up}}\\
\hline
SciBERT-Lin&allenai/scibert-scivocab-uncased& 2&AdamW &5e-5 &Linear for 10\% Training data\\
DeBERTa-NLI&microsoft/deberta-base-mnli& 2&AdamW &5e-5 &Linear for 10\% Training data\\
Siamese-Sim&sentence-transformers/all-MiniLM-L12-v2& 2&AdamW &2e-5 &Linear for 10\% Training data\\
T5-Sim&google/t5-base& 2&AdaFactor &1e-3 &-\\
T5-Gen&google/t5-base& 2&AdaFactor &1e-3 &-\\
\hline
\end{tabular}
\label{tab1}
\end{center}
\end{table*}

At Bosch, we build the Req2Spec tool for large scale automated software requirements analysis \cite{b20}. The goal of Req2Spec is to reduce the effort to manually validate the quality of requirements on aspects like consistency, vacuity, completeness, etc. This can especially be useful in large and complex projects, where the number of requirements can be substantial and manual assessment can be time-consuming and error-prone. It achieves this by translating natural language requirements into formal specifications using a series of NLP components. One of the key NLP components is a pattern classification model, that classifies a natural language requirement into one of the pre-defined restricted grammar categories from the formal specification pattern system described in \cite{b21}. The motivation of our paper is to tackle the low resource classification challenge of limited annotated data in industries and experiment with approaches such that pre-trained models are able to learn even with few annotated samples per category.

The remainder of this paper is organized as follows. Section 2 provides an overview of the experiment setup focusing on the data and architectures of the proposed approaches. Section 3 analyzes the results of the proposed approaches. Section 4 concludes the paper and proposes future work.

\section{Experiment Setup}

\subsection{Data}

We considered 3098 BOSCH automotive domain software requirements corresponding to the 3 most common pattern categories:

\begin{itemize}
\item \textbf{It is always the case that EXPR holds} (Total Count: 424)
\\E.g.: \textit{The window roll down time is 2 seconds}.
\item \textbf{It is always the case that if EXPR holds, then EXPR holds as well} (Total Count: 795)
\\E.g.: \textit{If ignition is on, then fuel indicator is active}.
\item \textbf{It is always the case that If EXPR holds, then EXPR holds after at most DURATION} (Total Count: 1879)
\\E.g.: \textit{If ignition is on, then the wiper movement mode is enabled within 0.2 seconds}.
\end{itemize}
The training set for each pattern was 15 requirements and the experiments were repeated with 3 random seeds for training data selection. Both the requirements and patterns text was lower cased to maintain consistency across the different models. Further, we also trained the models with 50 requirements (35 additional requirements along with the 15 previously chosen) per pattern to analyze how the performance of the different approaches scales with incremental data in the few-shot setting. In both cases, each requirement was augmented to generate an additional 50 unique samples with WordNet based synonym replacement (where 30\% of words in the requirement were randomly chosen for replacement) using the nlpaug library \cite{b22}. 

\subsection{Proposed Approaches}

We tackle the task of pattern classification by reformulating the problem as 5 different NLP tasks to test their performance in a FSL setting. The encoder for each of the approaches was chosen based on the state-of-the-art performance achieved on the task. Fig. 1 shows an overview of the different approaches with a sample requirement. Table \ref{tab1} has the details of the hyper parameter settings used for fine-tuning the different approaches. The 5 different approaches are explained below:

\begin{enumerate}
\item \textbf{SciBERT encoder + Linear classification head (SciBERT-Lin)}: SciBERT \cite{b23} is a pre-trained deep learning model that is specifically designed for scientific text data. It is based on the BERT architecture, one of the most widely used deep learning models for NLP. SciBERT is pre-trained on a large corpus of scientific text, including research articles, patent applications, and scientific books. One of the main benefits of SciBERT is that since it is specifically designed for scientific text, which is known to be different from general text in terms of vocabulary and writing style. By pre-training on scientific text, SciBERT is able to learn rich representations of scientific language and capture the specific features and patterns of this type of data, which can lead to improved performance on NLP tasks in the scientific domain. This has motivated us to use the SciBERT encoder with a standard 3 neuron Linear classification head (corresponding to the 3 categories) for automotive domain software requirements, with the input and target constructed in the following way:

\begin{itemize}
    \item Training: $<$Requirement$>$ is fed as the input and the target is 1 for the linear layer output neuron corresponding to the correct pattern and 0 for the other two incorrect patterns.
    \item Inference: $<$Requirement$>$ is fed as the input and the pattern producing the highest output probability (computed with softmax activation on the linear layer) is picked as the predicted class.
\end{itemize}

\item \textbf{DeBERTa encoder + Natural Language Inference head (DeBERTa-NLI)}: Natural Language Inference (NLI) can be used as a way to do text classification. In NLI, the task is to determine the relationship between two input sentences, such as whether one sentence entails, contradicts, or is neutral with respect to the other. This relationship can be thought of as a form of text classification, where the goal is to classify the relationship between two sentences into a set of predefined categories.

By using NLI as a form of text classification, we can leverage pre-trained models that have been trained on large amounts of NLI data to perform text classification tasks with limited labeled data. For example, a pre-trained NLI model could be fine-tuned on a few labeled examples of a target text classification task to perform well in a few-shot learning setting. We utilize the DeBERTa model as an encoder for the NLI task, since it has shown improvements over both BERT and RoBERTa models by introducing novel techniques such as disentangled attention mechanism, an enhanced mask decoder and virtual adversarial training method for fine-tuning. We reformulate the classification problem as an Entailment vs Contradiction task, where the input and target to the model is constructed in the following way:
\begin{itemize}
    \item Training: With the correct pattern, 1 input is constructed as $<$Pattern,Requirement$>$ with a target of \textit{Entailment}, and with the incorrect patterns 2 separate $<$Pattern,Requirement$>$ inputs are constructed with the targets of \textit{Contradiction}.
    \item Inference: 3 $<$Pattern,Requirement$>$ inputs are constructed with each of the patterns, and the pattern producing the highest \textit{Entailment} probability is picked as the predicted class.
\end{itemize}

\begin{table*}[htbp]
\caption{Few-shot learning performance of the algorithms with 15/50 training seed samples per class}
\begin{center}
\begin{tabular}{|c|c|c|c|c|c|c|}
\hline
\textbf{Algorithm}&\multicolumn{6}{|c|}{\textbf{Metric}} \\
\cline{2-7} 
\textbf{} & \textbf{\textit{Macro F-1 score}}& $\Delta_{MacroF-1}$& \textbf{\textit{Weighted F-1 score}}& $\Delta_{WeightedF-1}$& \textbf{\textit{Accuracy}}& $\Delta_{Accuracy}$\\
\hline
SciBERT-Lin& 83.66 / 85.33& 1.67&87.33 / 89& 1.67 &87 / 88.66& 1.66\\
DeBERTa-NLI& 84 / 85& 1&87 / 88.33& 1.33 &87 / 88& 1\\
Siamese-Sim& 84.66 / 85.66& 1&85 / 89& 4& 84.66 / 88.66& 4\\
T5-Sim& 79.33 / 85.66& 6.33 &83 / 88.66& 5.66 &82.66 / 88.33& 5.67\\
T5-Gen& 82 / 86& 4&85.66 / 89& 3.34 & 85.66 / 89& 3.34 \\
\hline
\multicolumn{6}{l}{$^{\mathrm{a}}$All values are in \% averaged across 3 random seeds $^{\mathrm{b}}$\(\Delta \) is the difference between values at 50 and 15 training samples}
\end{tabular}
\label{tab2}
\end{center}
\end{table*}

\item \textbf{Siamese network with MiniLM encoder + Semantic Textual Similarity head (Siamese-Sim)}: Contrastive learning (CL) is a type of machine learning technique that learns to distinguish between similar and dissimilar samples in a dataset. CL is commonly used in computer vision and NLP applications when labeled data is scarce, since it focuses on learning a similarity function between samples rather than learning a standard input-target mapping function. A model is trained to learn a representation of each input sample that maximizes the similarity between positive pairs (i.e., have the same class label or belong to the same cluster) and minimizes the similarity between negative pairs. The training process involves selecting an anchor sentence from each class, which is a reference sentence that is used to learn a representation of each sample that captures its meaning and relationship to other samples. The model is trained to identify whether other sentences are similar or dissimilar to the anchor sentence.

One common method for CL is to use a Siamese network architecture, where two identical sub-networks are trained with shared weights to compare two inputs in parallel and generate a similarity score. By training the Siamese network on a large corpus of text data, the network can learn to capture semantic and syntactic features of text that are useful for text classification. In our approach, we utilize MiniLM as an encoder in the Siamese network \cite{b24}, which is a deep learning model that is designed to be smaller and faster than other transformer models like BERT and T5. MiniLM is designed to be a more efficient alternative to large transformer models, while still maintaining a high level of performance on NLP tasks. It is trained on a similar architecture as BERT, but with a smaller model size and reduced computational resources. This opens up new possibilities for NLP applications in fields such as few-shot learning.

We reformulate the classification problem as a Semantic Textual Similarity task, where the input and target to the model is constructed in the following way:
\begin{itemize}
    \item Training: With the correct pattern, 1 input is constructed with the $<$Pattern$>$ and $<$Requirement$>$ being fed in parallel to the two towers of the Siamese network with a target of \textit{1}, and similarly with the incorrect patterns 2 separate inputs are constructed with targets of \textit{0}. Each pattern plays the role of an anchor sentence.
    \item Inference: 3 pairs of inputs are constructed with each $<$Pattern$>$ and $<$Requirement$>$ to be fed to the two towers of the Siamese network in parallel, and the pattern producing the highest cosine similarity score with the requirement is picked as the predicted class. Each pattern plays the role of an anchor sentence.
\end{itemize}
\item \textbf{T5 encoder + Semantic Textual Similarity target (T5-Sim)}: T5 is one of the state-of-the-art transformer model that is based on a text-to-text framework and can be used for a wide range of NLP tasks. T5 is trained on a diverse range of tasks, with a simple text-based input/output format, so that the model can be fine-tuned for a new task simply by providing it with examples of the input/output text pairs. It is trained with a multi-task learning framework that enables it to perform well on a wide range of NLP tasks, including text classification, natural language generation, and machine translation. T5 can be fine-tuned on limited labeled data for specific NLP tasks, leveraging the knowledge learned from pre-training to achieve good performance with limited resources. We reformulate the classification problem as a Semantic Textual Similarity task, where the input and target to the model is constructed in the following way:
\begin{itemize}
    \item Training: With the correct pattern, 1 input is constructed as $<$Pattern,Requirement$>$ with a target of \textit{5}, and with the incorrect patterns 2 separate $<$Pattern,Requirement$>$ inputs are constructed with the targets of \textit{1}. The targets of 1 and 5 are chosen since T5 was also pre-trained with targets between 1 and 5 for the semantic textual similarity task.
    \item Inference: 3 $<$Pattern,Requirement$>$ inputs are constructed with each of the patterns, and the pattern producing the output token \textit{5} is picked as the predicted class. Although rare, since the model generates output text, it can happen that it either produces the output token \textit{5} for more than one pattern or for none of the patterns. In such a case, the pattern producing the smallest probability for the output token \textit{1} is picked as the predicted class.
\end{itemize}

\item \textbf{T5 encoder + Classification Generative Text target (T5-Gen)} : Instead of fine-tuning the T5 model to generate semantic textual similarity scores of \textit{5} and \textit{1}, we reformulate the classification problem as a pattern text generation task, where the model is given the requirement as an input and is expected to generate the correct pattern text as the output. The input and target to the model is constructed in the following way:

\begin{itemize}
    \item Training: $<$Requirement$>$ is fed as the input and the target is the correct $<$Pattern$>$ text.
    \item Inference: $<$Requirement$>$ is fed as the input and the output pattern produced by the model is picked as the predicted class. Although rare, since the model generates output text, it can happen that it either produces few extra or lesser output tokens for the pattern text. In such a case, the pattern with the smallest Levenshtein distance to the generated output text is picked as the predicted class.
\end{itemize}

\end{enumerate}

\section{Results}
The proposed approaches can be compared on two aspects:
\begin{enumerate}
    \item \textbf{A1}: FSL performance with 15 training seed samples per category.
    \item \textbf{A2}: FSL performance improvement with supplying an extra 35 training seed samples per category, which will give us insights into how the performance of these models can scale with additional samples.
\end{enumerate}

The first observation that can be made from Table \ref{tab2} is that for A1, the difference of the Macro F-1 score, Weighted F-1 score and Accuracy between the best performing and worse performing algorithm for each metric is 5.33 (Siamese-Sim vs. T5-Sim), 4.33 (SciBERT-Lin vs. T5-Sim) and 4.34 (SciBERT-Lin/DeBERTa-NLI vs. T5-Sim) respectively. It is interesting to observe for A2 that the same differences between the best and worst performing models reduces to 0.66, 0.67 and 1. We believe that the choice of algorithm plays a more crucial role when the number of training samples is severely constrained, but the performance becomes algorithm agnostic as the number of training samples increases. This can be attributed to the fact that they are all state-of-the-art pre-trained models which can adapt and converge to a similar performance.

Further, it can be seen from the $\Delta$ values that for A2, both the T5 based models showed large improvement across the metrics. We believe that the reason for this is that since T5 is a unified text-to-text model where it pre-trained every task as a text generation problem, supplying additional samples aids in fine-tuning towards a particular domain specific task. Siamese-Sim also showed considerable improvement in terms of Weighted F-1 score and Accuracy scores which we believe is due to the fact that the Siamese based architecture utilizes a metric (cosine similarity) based learning function for FSL which improves as more samples are provided.

In terms of learning from the input-target pairs, the approaches with DeBERTa-NLI, Siamese-Sim and T5-Sim explicitly require additional negative samples with each incorrect pattern for every requirement whereas SciBERT-Lin and T5-Gen rely only on a single requirement sample to learn the classification patterns. 

\section{Conclusion and Future Work}

In this paper we have proposed and evaluated 5 different methodologies on approaching a requirements classification task with FSL. While there is no one-size-fits-all model and architecture as the performance can vary depending the domain, quantity and quality of the available data, etc., we have provided a large-scale comparison using various state-of-the-art algorithms with real software project requirements of the automotive domain industry at BOSCH. The future directions of this work can involve fine-tuning the language models with domain specific data if available and adding domain specific vocabulary to the various pre-trained models. We also plan to apply the proposed approaches in other NLP tasks applicable to requirements engineering such as entity-relation classification, requirement scope categorization and entity resolution.

\vspace{12pt}

\end{document}